# Building Benchmarks from the Ground Up: Community-Centered Evaluation of LLMs in Healthcare Chatbot Settings

Hamna Hamna[*] [†]
Microsoft Research India
Bengaluru, India
hamnaabid@gmail.com

Gayatri Bhat[*]
Karya
Bengaluru, India
gayatri.bhat@karya.in

Sourabrata Mukherjee
Microsoft Research India
Bengaluru, India
t-somukherje@microsoft.com

Faisal Lalani
Collective Intelligence Project
New York, US
faisal@cip.org

Evan Hadfield
Collective Intelligence Project
New York, US
evan@cip.org

Divya Siddarth
Collective Intelligence Project
New York, US
divya@cip.org

Kalika Bali
Microsoft Research India
Bengaluru, India
kalikab@microsoft.com

Sunayana Sitaram[†]
Microsoft Research India
Bengaluru, India
Sunayana.Sitaram@microsoft.com

## Abstract

Large Language Models (LLMs) are typically evaluated through general or domain-specific benchmarks testing capabilities that often lack grounding in the lived realities of end users. Critical domains such as healthcare require evaluations that extend beyond artificial or simulated tasks to reflect the everyday needs, cultural practices, and nuanced contexts of communities. We propose Samiksha, a community-driven evaluation pipeline co-created with civil-society organizations (CSOs) and community members. Our approach enables scalable, automated benchmarking through a culturally aware, community-driven pipeline in which community feedback informs what to evaluate, how the benchmark is built, and how outputs are scored. We demonstrate this approach in the health domain in India. Our analysis highlights how current multilingual LLMs address nuanced community health queries, while also offering a scalable pathway for contextually grounded and inclusive LLM evaluation.

[*]Equal contribution
[†]Corresponding author

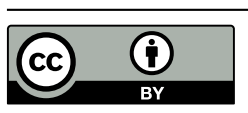

## 1 Introduction

Large language models (LLMs) are rapidly being integrated into high-impact socially relevant domains including agriculture, finance, healthcare, and law [18, 52, 58]. These systems are increasingly being adopted as tools for decision-making, information access, and public service delivery [56]. Yet their expanding reach has also drawn attention to broader ethical and social concerns, especially questions of adequacy for diverse audiences. Past research shows that LLMs often underperform for non-Western users, reflecting cultural misalignment and under-representation that surface as insensitivity, bias, and exclusion [4, 30]. These shortcomings not only undermine user trust, but also limit meaningful adoption of AI across global contexts. Addressing such issues requires evaluation methods that go beyond technical accuracy. Evaluations must ensure that responses are relatable, understandable, and practical for community members, reflecting their resources, cultural norms, and everyday challenges.

Benchmarking has become the dominant paradigm for evaluation, including efforts to assess cultural and domain-specific performance and cross-cultural representation [9, 51, 69]. However, today's benchmarks fall short: they often use translated or artificially created data [40] that overlook real user needs and elevate institutional priorities above those of the communities most affected. In contrast, a culturally grounded benchmark reflects local languages, values, and contexts; ensures representation across diverse identities; and employs community-validated criteria to evaluate models fairly and meaningfully. This gap between the promise of evaluation and the reality of existing benchmarks leads us to our central question: *How can we design scalable, culturally grounded, community-centered benchmarks to evaluate LLM-based systems?*

To address this, we propose Samiksha [1], an end-to-end evaluation pipeline that is guided by inputs from Civil-Society Organizations (CSOs) and data workers, who serve as end users. Our bottom-up approach combines the domain-specific expertise of CSOs with the lived experiences of Indian users. We conduct domain-specific evaluations of chatbots by eliciting community inputs in two complementary ways. First, we conduct focused interviews with CSOs in the relevant domains to capture their perspectives on common user query topics and evaluation requirements. Second, we translate these insights into task instructions for paid data workers, who create the benchmark query dataset and evaluate chatbot responses. These tasks are designed to let data workers draw upon their own lived experiences, interests, and concerns, thereby grounding the benchmarking work in their community realities. This integration of focused insights from CSOs with dispersed contributions from existing or potential chatbot users ensures that community concerns remain central throughout the process. This approach also channels expertise and engagement from multiple stakeholders, resulting in a benchmark that is both contextually grounded and adaptable to other domains and regions. The novelty of our work lies in this co-creation process and in the systematic integration of community feedback at every stage. CSOs were selected for their healthcare domain knowledge and regional familiarity, and data workers for their linguistic and cultural expertise, ensuring that the resulting benchmark is both representative and ecologically valid. We illustrate this pipeline through a case study in the healthcare domain in India.

Using the data points created through the above process, we create a benchmark covering several healthcare topics in three Indian languages: Hindi, Kannada and Malayalam. We conduct a fine-grained evaluation of three LLMs using both human annotators and LLM-as-judge methods [73], utilizing rubrics created in consultation with CSOs. Each model is evaluated along multiple dimensions: clarity & fluency, helpfulness & relevance, completeness & conciseness, and accuracy. Human evaluators contribute judgments shaped by cultural nuance, attending to local language use, social norms, and context-specific expectations of communication. In parallel, LLM-as-judge methods are employed to enhance efficiency, ensure consistency, and provide cross-validation of human ratings. We also analyze the evaluation behavior of humans and LLM-as-judges, finding that automated judges produce extremely stable yet compressed score distributions, while human raters exhibit broader variance and stronger sensitivity to linguistic nuance, completeness, and subtle factual distinctions.

Our work makes the following contributions:

(1) We present Samiksha, an evaluation pipeline that integrates inputs from CSOs and community members, centering lived realities in what to evaluate, how to build the benchmark, and how to score outputs. We provide a generalizable template that other domains and regions can adapt to build inclusive, context-aware LLM evaluations.
(2) Using this pipeline, we create a culturally grounded, multilingual benchmark spanning three Indian languages and provide empirical findings on three state-of-the-art LLMs.
(3) Through a mixed-methods analysis, we derive lessons learned and actionable recommendations from our community-centered evaluation study.

[1] The word "Samiksha" is of Sanskrit origin and means careful analysis or thorough investigation.

## 2 RELATED WORK

### 2.1 Current Landscape of LLM Evaluation

Large language models (LLMs) have been evaluated in diverse settings, such as multicultural [28, 61], multilingual [3, 69], and domain-specific contexts [8, 29]. Traditionally, evaluation has relied on standardized benchmarks [9, 67], synthetic datasets [13, 40], and expert-curated tasks [37, 53]. While these controlled evaluation methods remain valuable for measuring technical capabilities and enabling comparisons between models, they create a significant gap between how LLMs are tested and how they are actually used in real-world contexts [30, 31, 49]. Their key limitations are threefold. First, benchmarks are often too generic, meaning that models optimized for benchmark performance may struggle in open-ended or culturally specific situations. Second, most LLMs are trained primarily on English-language data, causing them to inherit racial, gender, and cultural biases present in these datasets [27, 41, 64]. Third, these approaches lack ecological validity, as they fail to account for local dialects and cultural nuances that are essential for populations that are under-represented in training data and do not primarily speak English [24, 32]. As LLMs become global tools, evaluation methods must evolve to reflect the cultural and social realities in which they operate [10, 31, 59]. Chiu et al. [10] propose CulturalBench, a benchmark of human-written questions covering 45 world regions. Their results show that even state-of-the-art models (like GPT-4o) achieve low accuracy especially on underrepresented cultures, emphasizing that LLMs often fail on nuanced cultural queries that go beyond textbook knowledge.

Several efforts have focused on adapting existing models and creating new language models specifically designed for non-English languages [46, 62, 65]. Researchers have also developed localized benchmarks for different regions, including India-specific evaluations [15, 19, 26, 66], as well as culture-specific assessment frameworks [10, 33, 39]. However, many multilingual benchmarks are simply translations of English-language benchmarks which can lead to loss of linguistic nuance, cultural meaning, and contextual relevance [39, 69, 71]. These limitations are especially consequential in high-stakes domains such as healthcare, law, and education, where model outputs directly inform decisions that affect people's lives. In these settings, issues such as misinformation, inappropriate recommendations, or cultural insensitivity can cause serious harm[12, 30].

To address these risks, recent work has advocated for community-centered, domain-specific cultural datasets and benchmarks that reflect local practices, values, and needs [31]. In the healthcare domain, the Open Medical LLM Leaderboard [44] consolidates model performance across a range of medical QA tasks, including MedMCQA [48] and MedQA [25]. These tasks are primarily short-form, exam-style multiple-choice questions (MCQs). Similarly, the LLM-in-Worldwide-Medical-Exams Leaderboard [74] evaluates LLMs on real professional medical exam questions from 28 countries, such as the USMLE and national licensure exams, using a pass/fail scoring system. These large-scale medical QA benchmarks have been developed using expert-authored content, medical board exams, and real-world patient–doctor dialogues. For the Indian context, MedMCQA-Indic [47] provides a translated version of MedMCQA, maintaining its medical entrance exam-style MCQ format while adapting it to Indian languages. While these datasets offer strong topical coverage and clinical relevance, they often lack grounding in community-specific cultural contexts and lived experiences. This underscores the limitations of one-size-fits-all evaluation frameworks and highlights the need for specialized, context-aware approaches. AfriMed-QA [43] marks a step forward by incorporating multi-stakeholder input and African contextual grounding, but limits its scope to short-answer and MCQ formats. In contrast, MenstLLaMA [1] foregrounds India-specific cultural sensitivity, focusing exclusively on menstrual health. Building on these efforts, our work, Samiksha, proposes a multilingual, culturally contextualized benchmarking approach grounded in real community queries. These queries capture the multi-layered, conversational, and dilemma-driven nature of community interactions, shaped by stigma, family dynamics, myths, power relations, and local resource constraints. Direct community feedback, enabled through multi-stakeholder involvement in Samiksha, informs what to evaluate, how the benchmark is constructed, and how model outputs are scored. These dimensions are essential for the responsible deployment of LLMs in real-world public health ecosystems. Table 1 compares the key features of our work with existing benchmarks.

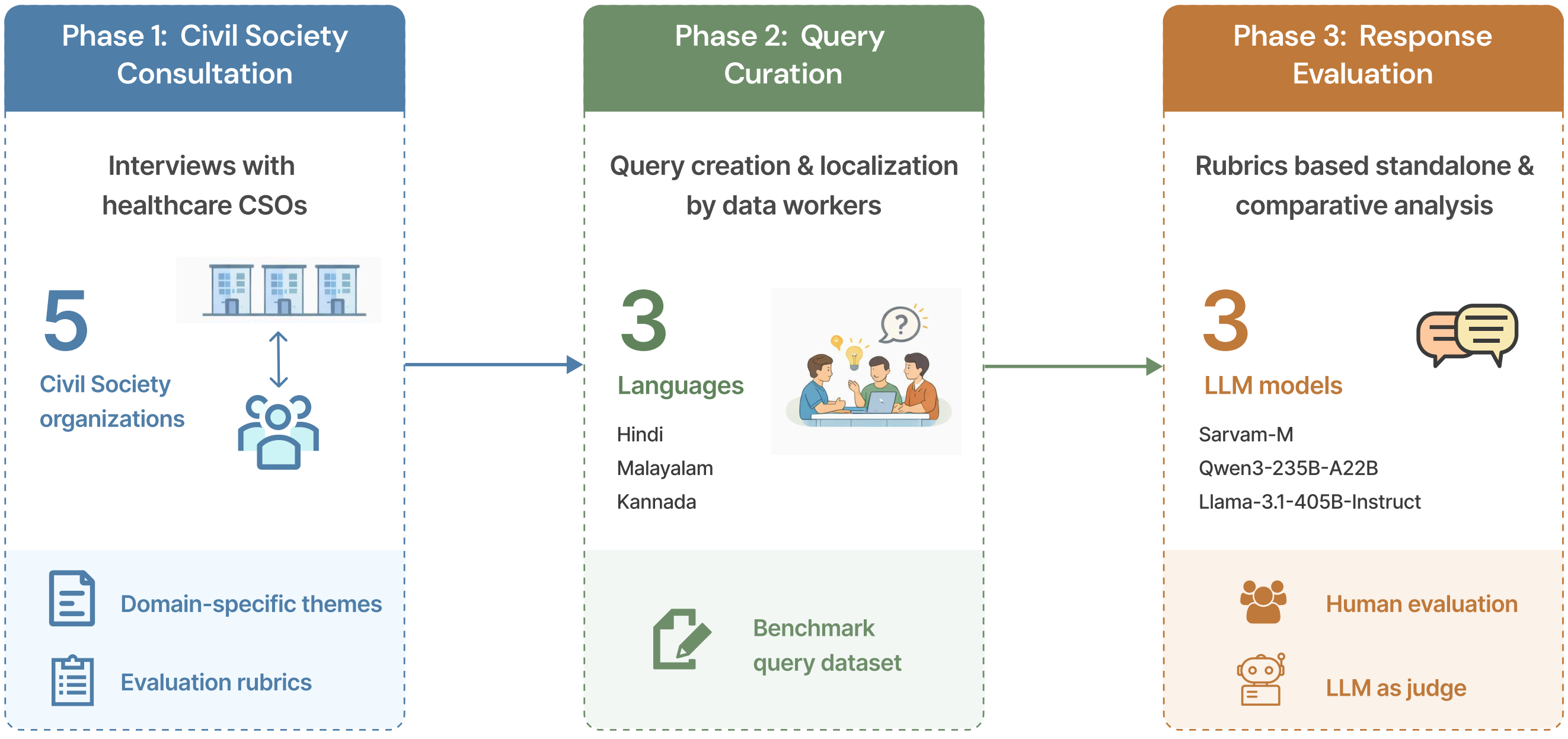

**Figure 1: The Samiksha Pipeline illustrates a structured, three-phase evaluation process designed to ground chatbot benchmarking in community realities. *Phase 1 – Civil Society Consultation:* Short, focused interviews are conducted with civil society organizations (CSOs) in relevant domains to inform benchmark and evaluation requirements. *Phase 2 – Query Curation:* Insights from CSOs are translated into task instructions for paid data workers, who create a benchmark query dataset. *Phase 3 – Response Evaluation:* Chatbot responses generated by three large language models are evaluated using a combination of human assessment and model-based judging. The data workers evaluate chatbot responses using their own lived experiences, interests, and concerns, ensuring that the assessment reflects authentic community needs and preferences.**

| Feature | Samiksha | AfriMed-QA[43] | MedMCQA[48] | MedDialog[72] | MedRedQA[42] | MenstLLaMA[1] |
|---|---|---|---|---|---|---|
| **Multi-stakeholders** | Yes | Yes | No | Yes | Yes | Yes |
| **Cultural questions** | Yes | Yes | No | Partial | Yes | Yes |
| **Long form answers** | Yes | Yes | No | Yes | Partial | No |
| **Multilingual** | Yes | Yes | No | Yes | No | No |
| **Indian culture** | Yes | No | No | No | No | Yes |
| **Human evaluations** | Yes | Yes | No | Yes | No | Yes |
| **Answer rationale** | Yes | Yes | No | Yes | Yes | No |

**Table 1: Comparison of Samiksha with existing medical QA datasets. "Multi-stakeholders" denotes the involvement of both end users and medical experts. "Multilingual" indicates the presence of languages other than English. "Answer rationale by users" refers to explanations or justifications authored by users.**

## 2.2 Community-Centered Approaches to AI Evaluation

There is a growing body of scholarship calling attention to globally inclusive and community-centered approaches to LLM evaluation [6, 22, 31, 50, 57]. One widely used method is crowdsourcing, which generally takes two forms. In the top-down model, corporate intermediaries such as Prolific or Amazon Mechanical Turk recruit community members as data workers, define the terms of participation, and dictate what counts as an acceptable submission, often leaving little space for feedback [38]. In contrast, the bottom-up model directly engages community members in data collection. This approach broadens opportunities for participation and empowers people not only to share cultural knowledge, artifacts, and expertise but also to shape the parameters of their own involvement [7, 11, 60]. STELA framework[6], for example, introduces a community-centered methodology to obtain norms for AI alignment. STELA conducts deliberative focus groups with underrepresented communities in the U.S. to define rules for chatbot behavior. Similarly, Qadri et al. [50] argue for "thick evaluations" of cultural representation, showing through workshops with South Asian communities that community-defined metrics offer richer insights than conventional quantitative ones. Our work operationalizes this idea in the context of LLMs, creating a scalable benchmark pipeline that embeds community perspectives into both data generation and evaluation.

The importance of co-creation, however, extends beyond method; it also depends on who participates. For example, Hall et al. [21] found that annotators living outside a region are more likely to view exaggerated or stereotypical depictions of that region as representative, while local annotators can draw on lived experience to provide more accurate assessments. This highlights the value of involving community members directly in AI evaluation. Complementary to this, expert-in-the-loop approaches bring in specific specialist stakeholders whose perspectives strengthen evaluation outcomes[52].

Building on these insights, we position Samiksha within broader responsible AI frameworks. We combine community data collection with domain-expert perspectives to co-create evaluation categories and judgments. This hybrid design differs from prior crowdsourced or expert-only evaluations by centering the lived realities of marginalized user groups while retaining structured input from practitioners familiar with the domain's operational constraints. Taken together, our co-creation process not only broadens participation but also yields a benchmark that reflects real information needs, cultural norms, and contextual constraints more effectively than existing evaluations.

# 3 Methodology

To assess whether an AI tool is genuinely useful to end users, it is essential to center civil society and community perspectives. Doing so, however, presents two key challenges. First, Civil-Society Organizations (CSOs) serve as vital stewards of community experience and domain expertise, yet limited resources often constrain their ability to participate intensively in AI evaluations. Second, while community members are best placed to assess a tool's usefulness in their everyday lives, they largely lack the opportunities or incentives to participate meaningfully in evaluations.

The Samiksha pipeline is designed expressly to address these challenges, and to embed civil-society and community perspectives into the entire AI evaluation process. The pipeline begins with focused CSO consultations and then translates insights from these consultations into large-scale evaluations through paid data work carried out by community members. This approach systematically combines high-level perspectives from CSOs, which offer broad coverage of community needs and priorities, with fine-grained inputs from individual community members that capture lived experiences. By integrating these complementary inputs, the pipeline enables a holistic and grounded understanding of community requirements in interactions with chatbots.

In addition to human evaluation, we also assess LLMs-as-judges to explore whether they can reliably approximate human judgments. This helps determine their suitability as scalable evaluators, identifies potential gaps in alignment with human preferences, and informs how automated evaluation might complement community-driven assessments.

The evaluation pipeline consists of three phases: civil society consultation, query curation, and response evaluation (see Figure 1). It is designed to be applicable across domains, geographies, and languages. In this study, we prototype the pipeline by evaluating LLM performance on healthcare-related queries posed by Indian users in three languages: Hindi, Kannada and Malayalam. This project was conducted between June and August 2025 as a collaboration among three partner organizations, involving five civil-society organizations (CSOs) and fifteen data workers.

## 3.1 Stakeholders

***Research Partners:*** The research team was comprised of three collaborating organizations:

1. **Karya** aims to create dignified digital earning and learning opportunities for low-opportunity communities globally, by leveraging the AI revolution. With expertise in developing multilingual and culturally grounded datasets, the organization led participant outreach and all data collection efforts for this project.
2. **Collective Intelligence Project** is dedicated to democratizing AI evaluation and governance by involving people from around the world in developing, refining, and setting standards for AI systems. The organization conceptualized this project's community-centered design and multi-stakeholder approach.
3. **Microsoft Research India** provided methodological guidance for pipeline design and execution.

***Participants:*** We engaged two main participant groups in this study:

(1) **CSOs:** Partnerships with CSOs go far beyond simple data gathering. Being both community-centered and technically equipped, these organizations are well positioned to provide actionable inputs for evaluation. Many of them engage directly with marginalized groups such as low-income, rural, or other under-represented communities whose voices are often missing from mainstream datasets and benchmarks. They also bring strong contextual expertise, helping interpret queries with the social, cultural, and linguistic nuances that an external team might miss. Furthermore, CSOs help filter and frame user concerns in ways that make public input more readily translatable into technical evaluation frameworks. Finally, because the Samiksha pipeline relies on persistent interactions with participants, collaboration with CSOs helps build long-term relationships with communities rather than being one-time data collection exercises.

(2) **Data workers:** Karya recruited a small cohort of data workers for this project (see Table 3 for demographics). As native speakers of the three languages included in this study, data workers contribute essential linguistic competence; equally important, they provide the perspectives of everyday Indian chatbot users whose lived experiences shape how they engage with AI systems. All members of the cohort use AI chatbots regularly for personal and professional purposes, positioning them to assess queries and responses not as abstract annotators, but as representative end users. Their involvement enables the evaluation to capture practical expectations that are often absent from expert-only assessments. The data work tasks are explicitly designed to surface these user-centered perspectives and preferences across both query curation and response evaluation (see Subsections 3.3 and 3.4). Each data worker was compensated at the rate of Rs. 10 per query creation or localization microtask and Rs. 20 per response evaluation microtask.

| CSO ID | Chatbot deployment | Languages supported |
|---|---|---|
| CSO-1 | Yes | Hindi, Marathi, Telugu |
| CSO-2 | Yes | Kannada, English |
| CSO-3 | Yes | Hindi, English |
| CSO-4 | No | Malayalam |
| CSO-5 | Yes | Hindi, Kannada, English |

**Table 2: Participating healthcare CSOs**

| | |
|---|---|
| **Number of participants** | 15 |
| **Age range** | 19–36 |
| **Gender** | Male: 4 ; Female: 11 |
| **Education** | High-school: 1 ; Undergraduate: 10 ; Graduate: 4 |

**Table 3: Participant details**

## 3.2 Phase 1: Civil Society Consultation

This first phase of the pipeline focuses on eliciting high-level evaluation inputs from CSOs. We begin by identifying CSOs operating in the healthcare domain and working closely with local communities in India, ideally in one or more of the languages included in the evaluation. We then conduct interviews with these CSOs to understand existing and potential trends in community LLM usage, risks and opportunities in domain-specific LLM usage, the types of queries users might pose to chatbots, and what safe and useful responses to these queries might or might not look like. Insights from these consultations feed directly into the next phase of the pipeline.

In our study, we shortlisted five healthcare CSOs that work closely with local communities, most of which have experience with chatbot-based interventions. Some had already deployed chatbots, while others were in the process of planning deployments. Initial contact was made via email, including a brief description of the research goals and a request for a virtual interview. Five CSOs agreed to participate (see Table 2), with each organization self-nominating a member to serve as its representative. The CSOs were not compensated for their time.

We conducted a series of semi structured interviews with representatives from the five CSOs. Most interviews were conducted in English, with some sessions blending English with Hindi or Malayalam, depending on participant preference. The interview protocol (Appendix A) which guided our conversations was designed to be flexible so that emerging and unexpected themes could be explored in greater detail. Two researchers conducted each interview, one leading the discussion and the other taking detailed notes, as the interviews were not audio recorded.

The main aim of these interviews and follow-ups was to capture regionally relevant user needs and concerns, as per each CSO's expertise. Interviewees shared perspectives on domain-specific chatbot use, their own interventions to ensure safe and effective chatbot use in the communities they serve, and broader visions for the future of healthcare chatbots. They described the styles and themes of queries users pose to experts or chatbots, providing examples from on-ground interactions or chatbot deployments. CSOs also articulated the qualities of good and poor LLM responses to healthcare queries, either as general criteria or via specific examples. Conversations with CSOs that had deployed chatbots included lessons learned from development and deployment; two CSOs also shared actual queries received by their chatbots (with informed user consent). Interviewees were encouraged to elaborate on particulars as needed. After each interview, we iteratively refined our questions, continuing until responses reached theoretical saturation [55].

We conducted thematic coding of interview notes, grouping recurring points into 15 themes capturing healthcare query topics identified by CSOs. Eight broader themes emerged through generalization of these topics. To guide data workers in the subsequent query creation phase, we synthesized topic-specific example queries in Indian English, reflecting diverse styles and perspectives. Table 4 lists the final query themes with definitions and example queries. We also synthesized CSO recommendations on evaluation rubrics, which are discussed in Subsection 3.4.

| Theme | Definition | Example queries |
|---|---|---|
| Access to community healthcare / primary healthcare | Questions about accessing healthcare via government schemes, facilities, professionals, costs, precautions against fraud, and health insurance. | I cannot speak freely with my gynaecologist because she tells my mother everything. How can I go to another doctor without making it awkward? |
| Managing injuries and infectious disease | Covers diagnosis, treatment, caregiving, symptoms, side-effects, prevention, and emergencies. | My brother has TB. What precautions should we take at home to ensure others don't get TB? |
| Managing chronic conditions | Concerns about long-term conditions (BP, diabetes, insomnia, snoring), family history, lifestyle changes, and medication. | My doctor has told me to add more iron to my diet. How to do this without non-veg? |
| Wellness habits | Covers exercise, diet, sleep, substance use, mental health, stamina, hair/skin care, and routine wellness. | I have low BP. Can I fast for Shivratri? |
| Reproductive health | Includes sexual wellbeing, contraception, family planning, and menstruation. | During periods, is it safe to have sex without a condom? Can the woman get pregnant? |
| Maternal health | Covers health during pregnancy, childbirth, and post-partum; risks and precautions. | Why shouldn't I eat mango or jackfruit when pregnant? |
| Children's health | Concerns about vaccinations, injuries, puberty, mental wellbeing, and emotional growth. | My son is always on the phone and doesn't go out to play. I think he is addicted, what to do? |
| Senior care | Caring for older family members: routines, medications, check-ups, and caregiver stress. | The doctor told my father to eat less salt but he won't stop eating salty snacks. What to do? |
| Everything else | Other healthcare-related questions not fitting specific themes. | |

**Table 4: Healthcare query themes collated from CSO consultation**

## 3.3 Phase 2: Query Curation

In the second phase of the pipeline, we develop a dataset of healthcare queries to evaluate LLM performance. To achieve both broad coverage of community concerns and fine-grained specificity of user needs, we expand on insights from civil society consultations through two tasks:

(1) **Query creation:** Data workers create healthcare queries based on the topics synthesized in the previous phase. To clearly frame the task, workers are asked to draft questions they would pose to healthcare experts such as doctors, nurses, or pharmacists. They are advised to produce realistic queries, include relevant contextual information, chain related questions where appropriate, and avoid sharing personal information. There is an option to skip topics.
(2) **Query localization:** Data workers produce localized variants of English queries by translating them into the target language and adding cultural context as appropriate. They are instructed not to substantially revise the underlying query. The source English queries are curated from chatbot logs shared by CSOs and undergo multiple rounds of filtering, including deduplication, removal of administrative, generic, or unintelligible queries, and removal of personally identifiable information (PII). (See example queries in Table 5.)

Karya recruited and trained data workers for both tasks. Training sessions situated the tasks within this AI evaluation project and introduced guidelines for query creation and localization. We recruited five native speakers each of Hindi, Kannada, and Malayalam. For the query creation task, each participant drafted approximately 50 queries in their native language evenly divided among all topics. For query localization, each participant localized approximately 50 queries from Indian English into their native language, carefully preserving the original intent and meaning. Upon task completion, Karya reviewed all outputs and provided specific feedback, and data workers made revisions as necessary. In all, the query creation and localization tasks yielded 270 and 260 queries per language (810 and 780 queries in total), respectively.

| |
|---|
| Can I drink coffee during TB treatment? |
| Is there a special mattress for sleeping during pregnancy? |
| Which home medicine should be given if a 7-year-old child is vomiting? |
| Is it necessary to make an appointment with a doctor to get the vaccine? |
| Under the Magare Karam scheme, on which days do pregnant and lactating mothers receive their check-ups? |
| Tell me about implants. |

**Table 5: Sample of healthcare queries selected for localization**

The queries created from scratch spanned a diverse range of factual, advice-seeking, and open-ended questions, with varying levels of contextual detail. Notably, some queries addressed social determinants of health, including socio-economic conditions, cultural beliefs and stigma, and gender norms that shape health behaviors and care-seeking decisions. Table 7 presents a sample of queries created through this process, with additional examples provided in Appendix B.

## 3.4 Phase 3: Response Evaluation

In the third pipeline phase, we generate and evaluate LLM responses to the curated query dataset.

We choose not to create gold-standard reference answers for the query set for both methodological and practical reasons. On one hand, developing high-quality reference answers in the healthcare domain requires substantial expert time and careful validation, which limits scalability even for a relatively small set of questions. This work cannot be reliably delegated to non-expert data workers. On the other hand, the questions generated through our community-driven process are inherently open ended and context dependent. They reflect lived experiences, locally relevant concerns, and diverse informational needs, and therefore do not lend themselves to a single correct answer in the traditional benchmarking sense. Introducing gold-standard responses would risk flattening this variability and reintroducing the institutional biases our approach seeks to avoid.

Instead, as described in the next section, we gather responses from candidate models and ask both experts and non-experts to evaluate them using a structured rubric. The rubric is developed by synthesizing civil-society guidance from Subsection 3.2, and prioritizes criteria such as clarity, reliability and practical usefulness given local constraints. The evaluation itself is conducted via paid data work, directly incorporating the perspectives of community members and end users. By focusing on whether a response is meaningful and actionable for community members, rather than on alignment with a predetermined canonical answer, our evaluation better reflects the realities of community information needs and locally grounded user preferences.

# 4 Evaluation and Analysis

## 4.1 Setup

To assess the performance of Large Language Models (LLMs) on our community-driven benchmark, we selected three state-of-the-art multilingual models: Sarvam-M [63], Qwen3-235B-A22B [70], and, Llama-3.1-405B-Instruct [17]. We generated model answers to all queries created from scratch (totalling 810 data points) in three Indian languages (Hindi, Malayalam, Kannada) by prompting each model with an identical instruction template specifying the role of "health expert", the target language, and a uniform word limit for all responses to prevent length bias [23]. We used the hyperparameters recommended in each model's technical documentation for decoding.

All model responses were evaluated by data workers, alongside experiments using LLMs as evaluators [34, 36]. The LLM-as-judge paradigm has gained traction in Machine Learning research because traditional metrics often fail to capture complexities, whereas LLMs can apply detailed rubrics to generate scores. While LLMs offer a scalable alternative for tasks that are otherwise costly and slow to complete manually, they frequently struggle to align with human judgments especially in multilingual and multicultural contexts [20, 54]. For this reason, we relied primarily on human evaluation, complemented by LLM-based scoring to explore scalability and consistency.

A cohort of 23 native-speaker data workers worked on the evaluation tasks, out of whom 12 had also worked on the query creation tasks. For the LLM-as-judge experiments, we prompted (see Section C) GPT-4o [45] and Sarvam-M [63] as judges to score and compare model answers. Note that Sarvam-M was also one of the models we evaluated, and we picked the same model as a judge to test for self-bias [36]. Both sets of evaluations produced categorical or ordinal judgments along with free-text rationales. All model outputs and judge responses were recorded for analysis.

## 4.2 Evaluation strategy

We used two complementary evaluation paradigms, standalone/absolute rating and comparative/pairwise-relative evaluation, using rubrics derived from CSO inputs.

***Standalone rating.*** Each response was rated independently on four dimensions:

i. **Clarity & Fluency:** Is the language clear, grammatically correct, and easy to understand?
ii. **Helpfulness & Relevance:** Does the answer directly address the question and provide useful information?
iii. **Accuracy (General Perception):** Based on general knowledge, does the information seem trustworthy and factually correct?
iv. **Completeness & Conciseness:** Does the answer provide sufficient detail without being overly verbose or repetitive?

For each dimension, annotators selected one of three options: *Yes*, *Somewhat*, or *No*. Annotators were provided with guidelines and examples explaining what each of the three options indicates in the context of each dimension. Annotators were asked to record a voice note for each rating, describing their thought process in making that selection.

LLMs-as-judges were prompted to provide the same judgments and rationales. These standalone ratings measured absolute values and fed into mean scores per model averaged across items, criteria, and languages.

***Comparative evaluation.*** We assessed relative performance via pairwise comparisons among the three candidate LLMs. We generated three responses ($A_1$, $A_2$, $A_3$) per question and constructed all possible pairs: $\{Q, A_1, A_2\}$, $\{Q, A_1, A_3\}$, and $\{Q, A_2, A_3\}$. Annotators were then asked, "*Which response do you think is better?*" and presented with three options: *Option A*, *Option B]*, and *[Not sure]*. Annotators were advised to choose the better answer as per their personal preference, and to record a detailed spoken explanation for this choice. No guidelines were provided on what constitutes a good or bad response, but coordinators shared guidelines and examples to convey what a detailed and informative spoken explanation could look like. Pairwise evaluations were aggregated into pairwise win-rates and into per-model win-share (win_rate) summaries under a consistent tie-handling policy [14]. Comparative judgments can surface subtle preferences and often reveal distinctions that are difficult to discern in absolute ratings, particularly when scores are very high [2].

| Questions | Translation |
|---|---|
| 6 വർഷം മുമ്പ് ഞാൻ എന്റെ മകളെ പ്രസവിച്ചത് സിസേറിയൻ ആയിരുന്നു. ഇനി ഒരു കുട്ടിക്ക് ശ്രമിക്കേണ്ട സമയം ആയെന്ന് എല്ലാവരും പറഞ്ഞു തുടങ്ങി. ഞങ്ങളും അതെ തീരുമാനത്തിലാണ്. എന്നാലും എനിക്ക് ഇനി സുഖ പ്രസവം ആവാൻ ആണ് ആഗ്രഹിക്കുന്നത്. അതിനു എന്തെങ്കിലും ബുദ്ധിമുട്ടുകൾ ഉണ്ടാവുമോ? എന്തൊക്കെയാണ് അതിനു വേണ്ടി മുൻകരുത്തേണ്ടത്? | I gave birth to my daughter 6 years ago via Cesarean. Everybody started saying that it's now time to try for another child. We are also deciding the same. Still I wish to have a normal delivery this time. Are there any challenges with that? What precautions should be taken? |
| ജീൻസ് പോലെയുള്ള വസ്ത്രങ്ങൾ തുടർച്ചയായി ഉപയോഗിക്കുന്ന പ്രത്യേകിച്ച് രാത്രി കാലങ്ങളിൽ ഉപയോഗിക്കുന്ന സ്ത്രീകൾക്ക് കുഞ്ഞുങ്ങൾ ഉണ്ടാകുവാൻ സാധ്യത കുറവാണ് എന്ന് ഒരു സുഹൃത്ത് പറയുന്നത് കേട്ടു. ഇത് സത്യമാണോ? | I heard a friend say that women who wear clothes like jeans continuously, especially at night, are less likely to have children. Is that true? |
| घर मे हर उम्र के लोग है सबका अलग अलग पसंद है और जब खाने की बात आती है तो मैं हमेशा असमंजस मे रहती हु\| क्युकी उसमे बुजुर्गों का अलग से प्रबंधन करना पड़ता है और कोई अच्छा उपाय क्या है जिससे इसको ठीक से मैनेज किया जा सके? | There are people of all ages at home, each with different preferences, and when it comes to food, I always find myself in a dilemma. Because elderly people need to be managed separately, what is a good solution to manage this properly? |
| घर पर बुजुर्ग महिलाएं कहती हैं कि बच्चों को पैदा होने के तुरंत बाद ही स्तनपान कराना चाहिए परन्तु हॉस्पिटल में डॉक्टर ऐसा करने से रोकते हैं, ऐसे में हमें क्या करना चाहिए? | Elderly women at home say that children should be breastfed immediately after birth, but doctors in the hospital prevent this; what should we do in this situation? |
| ಸೊಳ್ಳೆ ಕಡಿತದಿಂದ ನನಗೆ ಡೆಂಗ್ಯೂ ಬಂದಿದೆ, ಕೆಲವರು ಪಪ್ಪಾಯ ಎಲೆ ತಿಂದರೆ ಕಡಿಮೆಯಾಗುತ್ತದೆ ಎನ್ನುತ್ತಾರೆ, ಅದು ನಿಜಾನಾ ಅಥವಾ ಸುಳ್ಳಾ | I got dengue from a mosquito bite, some people say eating papaya leaves will help, is this true or false? |
| ನನಗೆ ಸಕ್ಕರೆ ಕಾಯಿಲೆ ಇದೆ ಆದ್ರೆ ಮಾತ್ರೆ ತಗೊಳೊಕೆ ಬೇಜಾರು ನಾನ್ ಮಾತ್ರೆ ತಗೊಳ್ದೆ ಹೇಗೆ ಸಕ್ಕರೆ ಕಾಯಿಲೆಯಿಂದ ದೂರ ಆಗ್ಬಹುದು? | I have diabetes but I don't feel like taking medication. How can I keep diabetes in check without it? |

**Table 6: User generated questions**

## 4.3 Analysis

Our analysis focused on how LLM judges compare with human judges (LLM-vs-human) and how LLM judges compare with each other (LLM-vs-LLM), using the two evaluation paradigms above and inter-evaluator correlation diagnostics.

***Standalone findings.*** Across standalone ratings, automated evaluators (GPT-4o and Sarvam-M as judges) produced compressed, near-ceiling mean scores on the study's ordinal scale, while mean scores from human annotators were lower and more dispersed. In other words, LLM judges were relatively lenient and less sensitive to moderate quality differences that humans were able to detect (see Figure 2a). Nevertheless, the *relative ordering* from standalone averages was broadly similar: Qwen3 tended to have the highest mean, Sarvam-M close behind, and Llama-3.1 lowest, indicating that automated judges and humans often agree on coarse rank order even when their absolute scales differ [69].

We further analyze distributional patterns for the three evaluators (Sarvam-M, GPT-4o, and Human) across the languages Hindi, Kannada, and Malayalam using violin plots that visualize the full score density for each criterion and model (Figure 3). Three consistent patterns emerge. First, automated evaluators, particularly GPT-4o, exhibit extremely narrow violins that indicate highly deterministic and low-variance scoring, while human ratings display substantially wider distributions, reflecting a 7.3× higher coefficient of variation (0.29 versus 0.04). Second, the variability changes across languages: Malayalam consistently produces the widest human violins, with variance approximately 1.48× higher than Hindi, suggesting increased linguistic ambiguity or lower evaluator familiarity. Third, variance also depends on the evaluation criterion. Completeness shows the highest variability (SD = 0.542), while clarity remains comparatively stable across evaluators and languages.

To establish that the differences between LLM and human evaluators are robust at the population level, we performed a set of complementary statistical analyses visualized in Figures 4a, 4c, and 4b. Figure 4a compares score distributions across four criteria using violin plots and shows that LLM evaluators consistently produce narrow, near-ceiling score densities while human evaluators display much broader, more variable distributions; Mann–Whitney U tests confirm these differences are highly significant for all criteria ($p < 0.001$). Figure 4c quantifies the magnitude of this divergence: LLM evaluators exhibit low variance and low coefficient of variation (CV ≈ 0.066), whereas humans show substantially higher variability (CV ≈ 0.273), amounting to a 4.15× consistency gap. Finally, Figure 4b presents 95% bootstrap confidence intervals for population-level mean estimates and shows that the intervals for LLM and human evaluators do not overlap for any criterion, providing strong evidence that true population means differ and that these effects are not sampling artifacts.

***Comparative findings.*** Comparative win-rate summaries reveal more nuanced judge-dependent rank shifts. Figure 2b shows that (i) `llama405b` is the least preferred model across judges, and (ii) the two leading models (`qwen3` and `sarvamm`) trade places depending on the judge: humans favor `qwen3` overall, GPT-4o favours `sarvamm`, while Sarvam-M-as-judge reports the highest win rate for `qwen3`. Thus, comparative evaluation exposes judge-dependent preferences that are not always visible in averaged standalone scores.

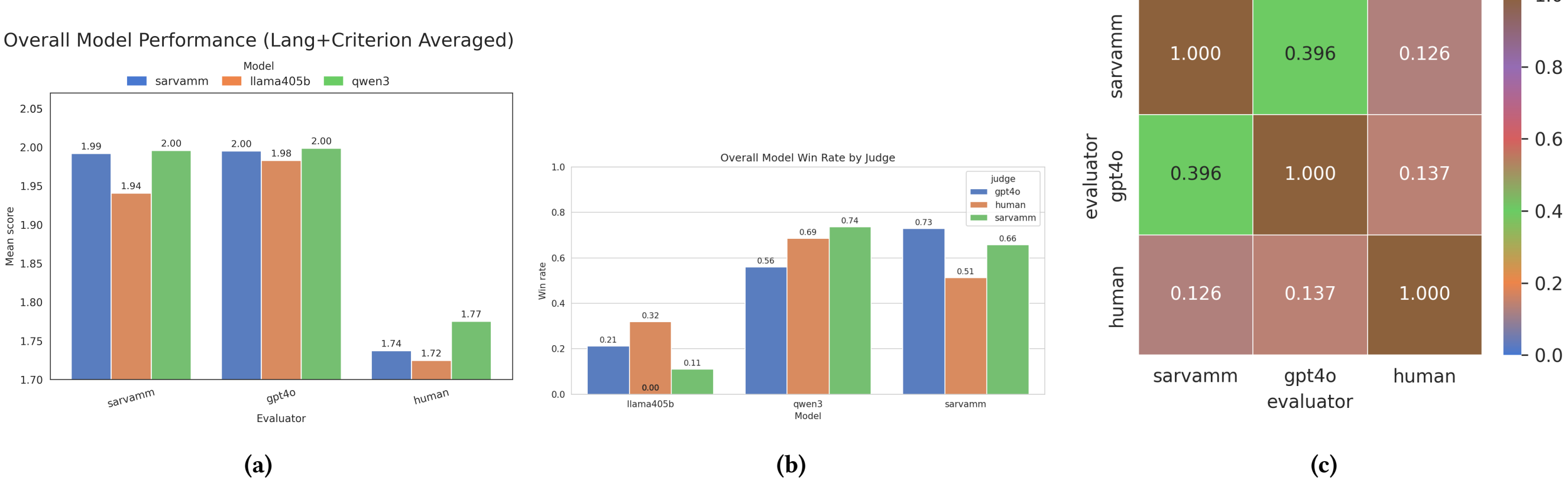

**Figure 2: Three-panel summary of benchmark evaluation: (a) absolute performance from standalone ratings, (b) aggregated pairwise comparative win rates by judge, and (c) inter-evaluator correlation (item-level means).**

A language-wise comparison of win rates further reveals how judge preferences interact with linguistic context. Across Hindi, Kannada, and Malayalam, all judges consistently rank `llama405b` as the weakest model, but the relative strengths of `qwen3` and `sarvamm` shift noticeably by language. In Hindi, humans strongly prefer `qwen3`, while GPT-4o leans toward `sarvamm`; Sarvam-M-as-judge also assigns its highest win rate to `qwen3`, amplifying its lead over `sarvamm`. For Kannada, this divergence becomes even sharper: human judges again strongly favor `qwen3`, whereas GPT-4o gives a clear advantage to `sarvamm`. Malayalam exhibits the same pattern of judge-dependent reversals, with GPT-4o consistently selecting `sarvamm` as the winner, while humans and Sarvam-M both slightly prefer `qwen3`.

***Inter-evaluator alignment.*** To quantify alignment we computed Pearson correlations over item-level mean scores across evaluators (sse Figure 2c). The LLM judges exhibit moderate mutual alignment (Pearson $r \approx 0.40$ between Sarvam-M and GPT-4o), while both LLM judges correlate only weakly with human annotators (Pearson $r \approx 0.13$). This indicates that automated judges capture a shared, model-centric view of quality that is distinct from the human evaluators' judgments: LLM-judges are more consistent with each other than with humans.

Building on these alignment differences, we compute Pearson correlations together with 95% confidence intervals (CIs) derived through Fisher's Z-transformation and visualize them in Figure 6. This representation reveals that human-GPT-4o correlations ranged from $r = 0.086$ (Helpfulness) to $r = 0.158$ (Accuracy), while human-Sarvam-M correlations were similarly weak, ranging from $r = 0.021$ (Helpfulness, $p \geq 0.05$, non-significant) to $r = 0.113$ (Accuracy). In contrast, inter-LLM correlations between GPT-4o and Sarvam-M were substantially stronger, with $r = 0.368$ (Accuracy) being the most notable, suggesting that LLMs exhibit greater consistency with each other than with human evaluators. The 95% confidence intervals highlight the statistical reliability of these differences: human-LLM pairs consistently showed CIs that were narrow and near zero (e.g., accuracy human-GPT-4o: CI $[0.090, 0.224]$, human-Sarvam-M: CI $[0.044, 0.180]$), indicating we can confidently generalize that these weak correlations represent genuine population-level disagreement rather than sampling artifacts. Moreover, criterion-specific variation emerged: Accuracy showed the strongest correlations across all pairs, while Helpfulness exhibited near-zero or negative correlations (e.g., human-Sarvam-M $r = 0.021$, $p \geq 0.05$), suggesting humans and LLMs assess helpfulness through fundamentally different lenses.

## 5 Discussion

### 5.1 Lessons from Civil Society Consultation

**Short and focused consultations** Recognizing that CSOs working closely with communities in high-stakes domains such as healthcare are often resource-constrained and balancing multiple urgent priorities, we designed the first phase of the evaluation pipeline to elicit their expertise through short and focused interviews. Our initial consultations and follow-up interactions with the five participating CSOs confirmed a strong willingness to contribute to the evaluation, alongside limited capacity to engage in more intensive activities such as generating query–response pairs or applying detailed evaluation rubrics to LLM outputs. In contrast, participation in a few hour–long sessions spread over the course of the project was feasible. By structuring these engagements carefully, we were able to gather rich, in-depth insights while respecting CSOs' time and capacity constraints.

**Loosely structured interviews** Although we initially planned highly structured interviews focused on eliciting prompt–response pairs (illustrative queries and gold-standard responses), we quickly pivoted to free-form discussions guided by a flexible interview protocol. This approach allowed each conversation to produce an unstructured stream of insights grounded in the healthcare challenges and technological interventions with which CSOs were most familiar. For example, one CSO articulated key motivations for India-focused healthcare chatbots, including stigma-free access to sensitive information (e.g., regarding sexual and reproductive health) and early support for users who may delay clinical care due to cost. Another CSO reflected on why their in-house chatbot failed to gain traction: healthcare workers preferred to consult colleagues or use general web search rather than navigate to a specialized chatbot. Across interviews, CSOs emphasized the need for chatbots that can handle dialectal variation, incorporate expert-in-the-loop review, and redirect users to nearby care when appropriate. While this loosely structured format required greater effort from the research team to distill and synthesize CSO inputs, the breadth of perspectives gathered informed design decisions throughout the pipeline. Even insights that did not directly shape this evaluation strengthened the motivation for the project and informed plans for a future, scaled-up evaluation.

**Figure 3: Evaluation consistency analysis across evaluators, languages, and criteria. Each cell in the $3 \times 3$ grid shows violin plots of score distributions for three evaluated models (Sarvam-M, LLaMA-405B, Qwen3) under a specific *evaluator–language* combination. Rows correspond to evaluators (Sarvam-M, GPT-4o, Human), and columns correspond to languages (Hindi, Kannada, Malayalam). Within each cell, violins depict kernel density estimates over four criteria: accuracy, clarity, completeness, and helpfulness, with embedded box plots marking medians and quartiles. Wider violins indicate higher score variance (less consistent evaluation), and narrower violins indicate concentrated, highly consistent ratings.**

**Synthesizing civil society inputs** CSOs shared illustrative user queries drawn from their work in areas such as high-risk pregnancy, tuberculosis, and menstrual health. We synthesized these

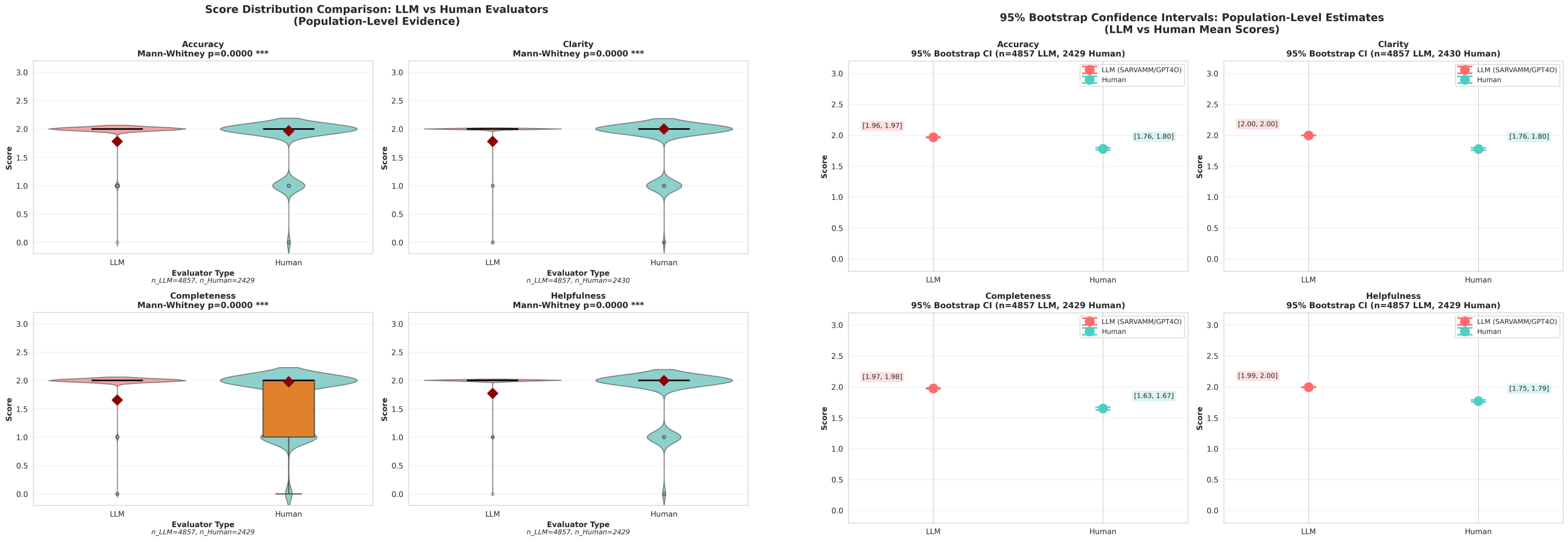

**(a) Score distribution differences (LLM vs. Human).**

**(b) Bootstrap 95% CIs for population-level means.**

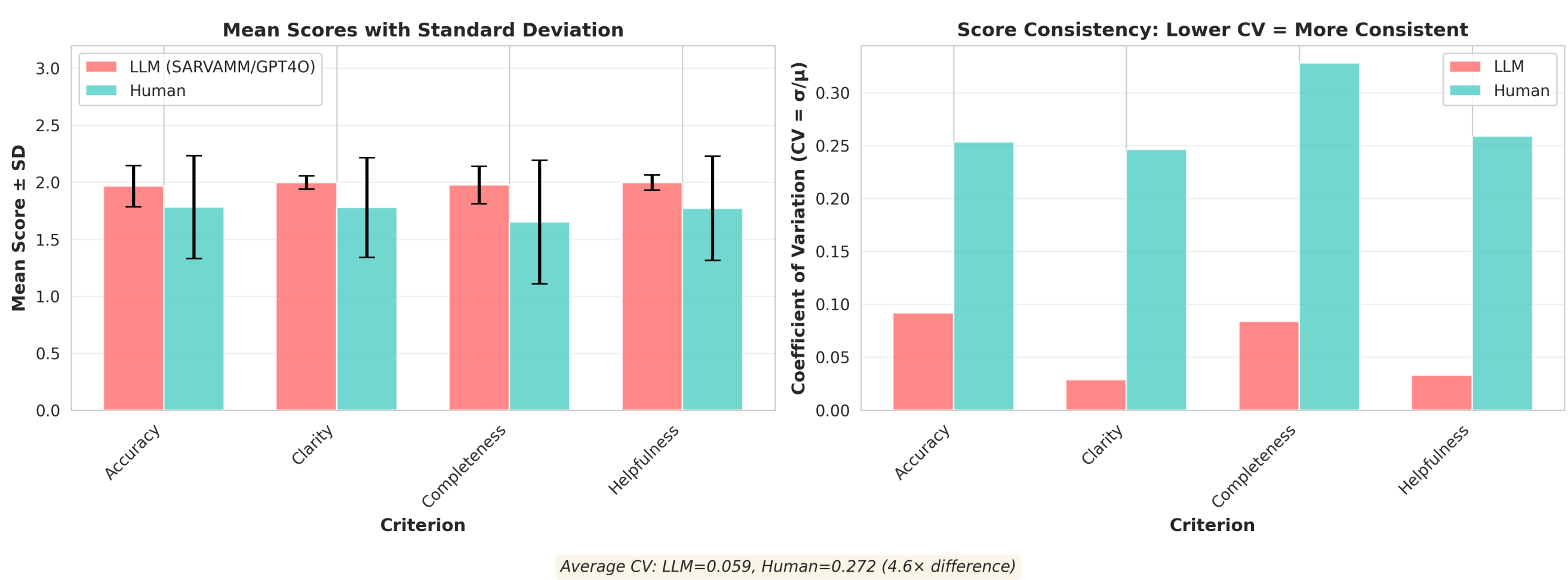

**(c) Variance and coefficient-of-variation comparison across criteria.**

**Figure 4: Comprehensive population-level statistical evidence comparing LLM and human evaluators across four evaluation criteria. Top row: (a) violin/box plots showing distributional differences between LLM and human scores; (b) bootstrap 95% confidence intervals demonstrating non-overlapping population means. Bottom row: (c) variance and coefficient-of-variation analysis quantifying the 4.15× higher variability in human judgments.**

inputs into a set of query topics with broad coverage and general relevance, including “maternal health”, “managing injuries and infectious diseases”, and “reproductive health”. We also collated a few example queries for each topic, in a range of styles. CSOs indicated that users—particularly women—may feel more comfortable raising sensitive questions with chatbots than with family members; accordingly, we retained sensitive categories such as “reproductive health”, while allowing data workers to opt out of any category during query creation. CSOs also articulated criteria for evaluating the effectiveness and shortcomings of LLM responses to healthcare queries. CSOs emphasized that responses should start with efforts to elicit missing contextual details, as users often provide incomplete information in initial queries. They highlighted the importance of using relatable and accessible language, offering informative responses without overwhelming users, maintaining a reassuring yet non-prescriptive tone, and remaining contextually grounded rather than prematurely authoritative or diagnostic. CSOs further underscored the need for socio-cultural sensitivity, noting that responses should account for users’ social and familial contexts rather than relying on narrowly individualistic framings. Considering the single-turn chatbot interactions being evaluated in this project, we distilled these insights into four criteria for response evaluation by non-experts.

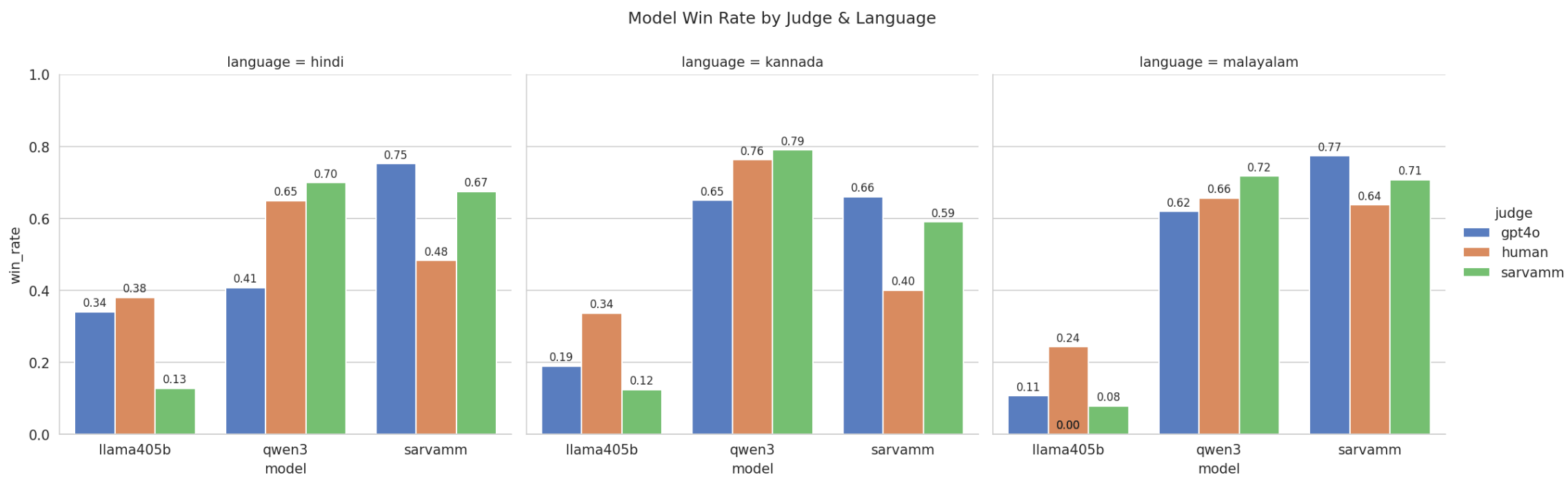

**Figure 5: Comparative winrates by language. Humans prefer Qwen3, while LLM judges favor Sarvamm, reflecting stylistic amplification.**

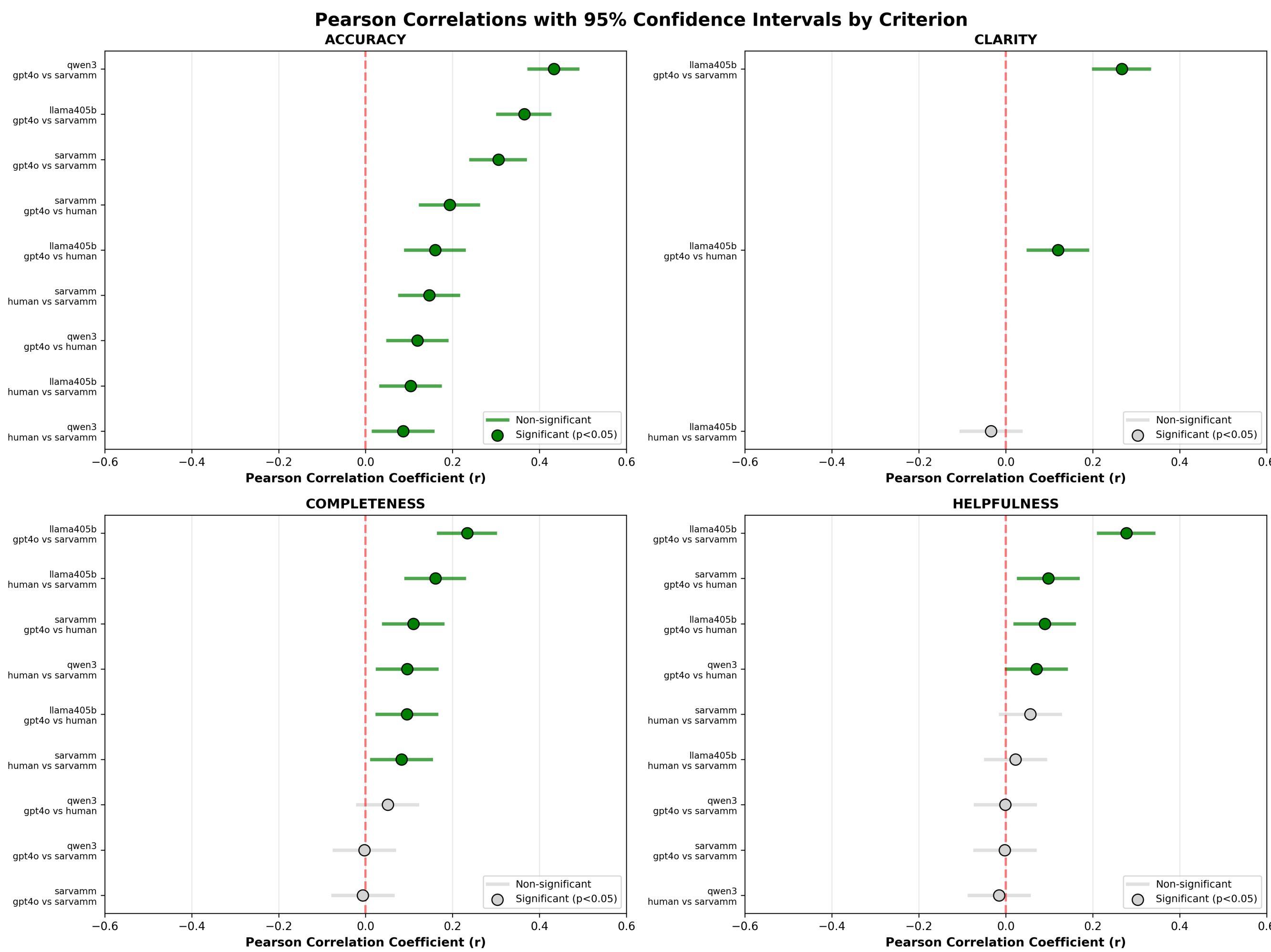

**Figure 6: Pearson correlations with 95% confidence intervals across all evaluator pairs and criteria. Each point shows the observed correlation and its uncertainty range, with green markers indicating statistically significant evaluator alignment ($p < 0.05$) and gray markers indicating non-significant associations whose CIs cross zero.**

**Layered perspectives** Because each CSO's recommendations are shaped by the projects they undertake and the issues they address, engaging with multiple organizations allowed us to develop a well-rounded understanding of the domain. CSOs sometimes offered differing guidance: one emphasized the utility of a reassuring but potentially slightly optimistic response (e.g., suggesting that medication will help while redirecting the user to a nearby healthcare facility), whereas another prioritized factually complete answers, supported by references. Insights from chatbot evaluations also highlighted variation in user preferences: CSOs told us that community healthcare workers, such as ASHA workers, tended to prefer concise responses, while patients and caregivers favored longer, more reassuring guidance. By consolidating these layered and occasionally conflicting perspectives, we developed an evaluation rubric that is a holistic lens for assessing LLM responses, rather than a rigid framework for categorizing "good" versus "bad" answers. CSO representatives consistently emphasized that their observations reflected their own experiences and were not a substitute for actual user input, reinforcing the importance of a multi-stakeholder, phased evaluation approach such as ours.

**Building a community of practice** As we plan to scale up our benchmarking and evaluation effort across more domains, geographies and languages, we envision CSO engagement as an ongoing, iterative process in which insights and feedback can be solicited regularly. To gather a depth and breadth of perspectives, we plan to combine various modes of engagement including surveys, interviews, round-table discussions, and document or data reviews. For instance, we will invite CSOs to review topic lists, example queries, user-developed queries and evaluation rubrics during the evaluation process. We will also invite CSOs to recommend experts who can participate in the response evaluation phase of the project. As the scope of these engagements expand, we will develop a compensation structure for participating CSOs and also tap their network of experts for paid data work. Wherever suitable, we will also explore the possibility of engaging "anchor CSOs" for deeper modes of collaboration depending on their capacity and resources. Based on mutual alignment of objectives and interest, an anchor CSO could facilitate interactions with other CSOs in its network, closely review design and execution, and ensure ongoing alignment with community needs.

## 5.2 Lessons from Query Curation

**Query Creation – Execution** The query creation task required data workers to draft 5–6 queries per assigned healthcare topic, a format markedly different from speech recording, transcription, and translation work they had previously undertaken. Initially, the cohort found this open-ended task challenging, as it lacked rigid guidelines. Project coordinators addressed this with a mix of written materials, training sessions, example queries, and individualized tutoring. Once oriented, the cohort developed a realistic set of 810 healthcare-related queries from scratch, evenly distributed across three languages, eight core topics, and one additional catch-all category. Data workers demonstrated a strong ability to produce high-quality, varied, and contextually grounded queries tailored to the assigned topics. Coordinators played a critical role in maintaining quality through continuous feedback and oversight, ensuring that queries were neither generic (e.g., "What issues come under [TOPIC]?") nor textbook-style (e.g., "What are two key practices that help ensure the health and safety of a mother during pregnancy?"). Guidance emphasized including sufficient context, using natural everyday language, and creating queries that are distinct from each other.

**Query Creation - Outcome** Users drafted a range of queries with both general outlooks (e.g., आज के दौर मे लड़कियों को समय से पूर्व मासिक क्यू आ जाते है? *Gloss: Why do girls get their periods so young these days?*) and situational outlooks (ഇന്നലെ ബൈക്ക് പെട്ടന്ന് ബ്രേക്ക് പിടിച്ചപ്പോൾ കാൽ പെട്ടന്ന് കുത്തണ്ടി വന്നു വീഴാതിരിക്കാൻ, ബൈക്ക് ചെരിഞ്ഞു കാലിലേക്ക് വന്നപ്പോൾ കാൽമുട്ടിൽ ഒരു വേദന അനുഭവപെട്ടായിരുന്നു, സ്കാൻ ചെയ്ത് നോക്കേണ്ട ആവശ്യമുണ്ടോ വേദന മാത്രം ഉള്ളുവെങ്കിൽ? *Gloss: Yesterday, I had to brake the bike suddenly and put my foot down to avoid falling. The bike leaned onto my leg and made my knee hurt. If pain is the only symptom, is a scan necessary?*). The queries also varied by intent: some sought information (e.g. ಭಾರತದಲ್ಲಿ ಸಾಂಕ್ರಾಮಿಕ ರೋಗಕ್ಕೇ ಅತಿ ಹೆಚ್ಚು ಬಡವರೇ ತುತ್ತಾಗುತ್ತಾರೆ ಏಕೆ? ಮತ್ತು ಹೇಗೆ ಇದನ್ನು ತಡೆಗಟ್ಟಬಹುದು? *Gloss: Why are the poor most affected by infectious diseases in India? And how can this be prevented?*), some fact-checked claims (e.g., പ്രഭാത വ്യായാമം ശീലമാക്കുന്നത് മൂലം മനസ് ശാന്തം ആകുമെന്ന് പറയുന്നത് ശെരിയാണോ? *Gloss: Is it true that the mind becomes calmer when morning exercise becomes a habit?*), and others requested advice (e.g., बुजुर्गों की देखभाल करने मे परेशानी बहुत आती है लेकिन मेरा मानना है अगर अगर सही रूटीन का पालन किया जाए तो ये उतना भी मुश्किल नहीं है फिर भी कुछ अच्छे उपाय की जानकारी चाहिए ताकि उनका ख्याल और अच्छे से रखा जा सके। *Gloss: Caring for seniors can be frustrating but I feel it isn't that difficult if there is a good routine, I still want some good ideas to take better care of them.*).

The queries created were multi-layered, reflecting how health-related concerns are deeply influenced by a user's family, community, and cultural values (see Table B). They went beyond straightforward medical questions to reveal the everyday dilemmas people face when balancing traditional beliefs, intergenerational advice, and modern medical practices. For example, childbirth choices after a previous Caesarean section reflect not only medical safety but also family expectations around "normal" (vaginal) delivery, while myths linking infertility to clothing choices illustrate how community myths shape perceptions of health. Queries about elder care highlight the challenges of multigenerational households, where food and lifestyle decisions must accommodate different needs. Similarly, dilemmas around breastfeeding illustrate the tension between family elders' authority and hospital protocols. Other queries point to reliance on indigenous remedies and local practices such as using papaya leaves for dengue, and skepticism toward dependence on modern medications, as seen in the hesitation to take diabetes treatment. Across all queries, recurring tensions emerge: traditional knowledge versus modern science, personal choice versus family authority, and evidence-based medicine versus superstition. Some queries also reveal gaps in rural healthcare access, limited awareness of government schemes, and confusion about reliable sources of information. Overall, these queries demonstrate that people's health concerns are rarely purely medical. Instead, they are embedded within social norms, cultural practices, and community narratives that shape how individuals seek care and make decisions.

**Query Creation - Recommendations** For this project, queries were created and validated on collaborative spreadsheets. To scale the process, we are designing integrated 'Query Creation' and 'Query Validation' workflows on Karya's proprietary data annotation platform, aimed at preserving this level of quality across larger evaluation efforts. Lessons from coordinators' oversight are being distilled into enhanced training materials to better prepare future cohorts. While topic-wise example queries helped participants navigate the open-ended task, they also introduced some bias. For instance, an example query about snakebite treatment prompted multiple queries on rat bites, and an example concerning pregnant women avoiding mango and jackfruit might have primed a worker to draft a query about papaya leaves. Planned revisions to task design include mechanisms to identify and manage near-duplicate queries. However, certain duplicates may reflect genuine shared community concerns such as multiple queries addressing loneliness among older adults and retaining these can enrich the dataset's representativeness. The current workflow ensures equal topic coverage but does not regulate query style. CSOs can provide insights into the real-world distribution of query topics and styles, which could inform dataset curation. Queries created from scratch often skew toward information-seeking, lower-urgency questions, rather than pressing or critical concerns that users encounter in practice. To balance this, it might useful to supplement these queries with examples drawn from domain-specific chatbot logs and online forums, which tend to capture more urgent, situational queries.

**Query Localization – Execution** The query localization task, as well as the validation of localized queries, readily translate into microtasks on Karya's annotation platform. Because all data workers recruited for this project had prior experience with other dataset creation tasks, they found the localization task straightforward to execute. The cohort localized 260 healthcare-related queries, extracted from the logs of a deployed chatbot, from English into Hindi, Kannada, and Malayalam. Data workers produced meaning-preserving localizations and researched domain-specific terminology where required. Nevertheless, accurately localizing certain queries proved challenging for non-experts, as preserving intent and meaning in queries such as "how to take the shadow pill" or "Which vaccines are given along with Penta?" requires specialized domain knowledge.

**Query Localization - Outcome** Queries extracted from real chatbot logs often originate from multi-turn interactions and may not function effectively as standalone queries. Compared to queries created from scratch in the query creation workflow, these queries contained less explicit detail and relied more on implicit context. In addition, many queries contained references to generic medicines and government schemes which might not translate cleanly across languages or geographies. As a result, even after extensive pre-processing to select the most suitable queries, data curated using this approach were not included in either human evaluation or LLM-as-judge evaluations.

**Query Localization – Recommendations** We nevertheless intend to continue using the query localization approach wherever suitable source data is available. In our expanded data collection process, we plan to treat English queries as "seed queries" and ask data workers to expand and revise them as needed, including adjusting the original intent and adding substantial context, to produce queries that are meaningful within their local settings. When datasets appropriate for localization are available, we recommend rigorous upfront filtering to remove administrative queries, duplicate or near-duplicate entries, content that is inappropriate or sensitive. Given that generating a large number of topic-specific queries from scratch can be resource-intensive, and that localization-derived queries can reflect authentic real-world information needs, this approach represents a valuable supplementary method for query curation.

## 5.3 Lessons from Response Evaluation

**Human Evaluation** Response evaluation is conducted as microtasks on Karya's annotation platform. We find that providing comprehensive guidelines, detailing the annotation rubric, rating scale, and including examples of detailed versus vague explanations is essential for ensuring consistency and clarity. Training remains critical: once data workers understand that they are serving as evaluators rather than being evaluated themselves, they are much more open to giving critical but balanced and nuanced feedback. Overall, data workers reported that comparative evaluation tasks are quicker and easier to complete than standalone evaluations. In feedback surveys, participants noted that assessing responses on accuracy was challenging, even when they were asked only for their general perception, but they were more confident evaluating criteria such as clarity, relevance, and conciseness. This underscores the need for a distinct rubric specifically designed to capture expert judgments on criteria including accuracy and safety, and a deeper ongoing engagement with CSOs and other domain experts.

**Human Evaluation vs. LLMs-as-Judges** While LLM-as-judge helps rank models at scale [16, 35, 68], our correlation, variance, and population-level analyses (see Section 4.3) paints a clear picture: automated evaluators (Sarvam-M and GPT-4o) agree with one another but diverge from human judgement: human judges strongly penalized missing information, factual errors (even if subtle), and language-specific weaknesses that LLM judges tended to overlook. This divergence is not only reflected in weak Pearson correlations with humans but also in score distributions and variance patterns: LLM judges exhibit highly compressed, near-ceiling scores with extremely low variance (CV $\approx 0.06$), whereas humans display far richer, more discriminative variability (CV $\approx 0.27$), including strong language- and criterion-specific effects. Our significance and confidence-interval analyses further confirm that these differences are population-level phenomena: distributional gaps between humans and LLMs are statistically extreme (Mann–Whitney $p < 10^{-100}$), and 95% bootstrap confidence intervals for population means do not overlap for any criterion. In the case of comparative ranking, LLM-judges deliver a reliable *relative* signal (regarding the better model) but suffer three important limitations in absolute quality assessment: (i) a compression or ceiling effect that reduces sensitivity to moderate differences, (ii) under-detection of completeness and nuanced factual inconsistencies, and (iii) lower language-specific sensitivity compared to human raters. Our study suggests that LLM-judges cannot be used as replacements for human evaluators in such contexts. Nonetheless, to support a benchmark that spans diverse languages and domains and testing tens to hundreds of models, we will need some automation. Purely human evaluation of tens of thousands of data points would be prohibitive in terms of time and cost. Our experiments suggest that a hybrid evaluation workflow in which LLM-judges perform high-throughput ranking and triage, and targeted human annotation validates and calibrates the most important or risky decisions, may be a solution.

## 6 Conclusion

LLM evaluation must move beyond English and Western-centric benchmarks to address tasks, domains, and contexts grounded in the lived realities of users worldwide. Central to this shift is embedding community perspectives directly into the evaluation process. Samiksha demonstrates that a community-driven pipeline that integrates civil society and user perspectives can surface culturally specific needs, shape what gets evaluated, and determine how outputs are scored. Demonstrated in the Indian context and initially in the healthcare domain, Samiksha offers a practical blueprint for more inclusive evaluation. The pipeline itself is domain-agnostic and can be adapted to other geographies with appropriate stakeholders and communication strategies, enabling broader applicability. Looking ahead, Samiksha provides a scalable template for extending evaluation across additional languages, contexts, and domains, paving the way for truly community-centered benchmarking of LLM technologies.

## Acknowledgments

We sincerely thank the community members who participated in this evaluation, as well as the civil society organizations (CSOs) ARMMAN, Eka Care, iLab, KHPT and KhushiBaby, whom we consulted, for their dedicated engagement and invaluable inputs throughout the project. We also extend our gratitude to Karya's operations team and project coordinators for their essential support. We are deeply appreciative of the time, thoughtful contributions, and insights shared by all participants.

We acknowledge the use of ChatGPT and GitHub Copilot in supporting parts of the code development and for minor language editing to enhance manuscript clarity. All content was reviewed and verified by the authors, who take full responsibility for the final version of the paper.

## Ethics Statement

We use the framework by Bender and Friedman [5] to discuss the ethical considerations for our work.

***Institutional Review:*** All aspects of this research were reviewed and approved by the Institutional Review Board of our organization and also approved by Karya.

***Compute/AI Resources:*** All experiments were performed on cluster of 8 MI300 192 GB AMD GPU and/or on 8 NVIDIA A100 80GB PCIe GPUs. API-based interactions with GPT models were facilitated through the Azure OpenAI service.

# A CSO interview protocol

1. General Introduction & Community Work
    - Please give us a brief overview of the CSO: its work/product, deployment location, and your role within the organization.
    - Which healthcare areas do you primarily focus on?
    - Who are your main target users?
    - From your experience, what are the key health-related needs and challenges you see in your community?
2. Community Experience with Digital Tools (before chatbots)
    - Have you used or introduced any digital tools or apps in your work so far? What has the experience been like?
    - How do community members usually engage with technology in health contexts?
3. About Chatbots (if applicable or exploratory)
    - What is the current status of your chatbot (if any)?
    - Who do you envision as the primary users of such a tool?
    - How do you see chatbot adoption in your community in the near future?

4. Common Health Queries
    - What are the most common health-related queries, and which of these can realistically be addressed by technology?
    - Whom do people usually turn to with their health questions?
    - Have you observed cases where reliance on AI tools/internet led to harmful or false information? If yes, what were they about?
    - Who are the typical users in such cases, and what kinds of queries do they bring?
    - Looking ahead, what questions might communities ask if they had access to a reliable chatbot? What barriers might prevent access?
5. Benchmarking and Evaluation
    - In your view, what would an ideal and ethical health chatbot look like?
    - In which situations might chatbot responses be misleading?
    - Simple Exercise: We will share a few hypothetical user queries. Please share your opinion on these queries, and highlight potential green flags and red flags in the expected chatbot responses.
6. Concluding Questions & Future Engagement
    - Would you be open to partnering with us by providing input on the quality of queries we collect?
    - Is there anything else you would like to highlight—such as challenges, support needs, or priority directions?
    - Could you share example queries or sample data to help us better contextualize community needs?

## B User generated queries

## C LLM Prompt Details for Evaluation (LLMs as Judges)

We employed standardized evaluation prompts to ensure consistent scoring across LLM judges.

**Standalone Prompt**

**Task:** Evaluate the answer to the question using four criteria. **Audience:** Internal evaluator. Do not browse. Rely only on general knowledge and the given text. **Language note:** The question/answer may be in Hindi, Malayalam, or Kannada. Judge clarity within the answer's language, but write explanations in concise English.

**Evaluation Criteria (choose exactly one rating per criterion):**

i. **Clarity & Fluency:** Is the language clear, grammatically correct, and easy to understand? Ratings: `Yes` | `Somewhat` | `No`
ii. **Helpfulness & Relevance:** Does the answer directly address the question and provide useful information? Ratings: `Yes` | `Somewhat` | `No`
iii. **Accuracy (General Perception):** Based on general knowledge, does the information seem trustworthy and factually correct? Ratings: `Yes` | `Somewhat` | `No`
iv. **Completeness & Conciseness:** Does the answer provide sufficient detail without being verbose or repetitive? Ratings: `Yes` | `Somewhat` | `No`

**Scoring Rule:** For each criterion, assign: `Yes = 2`, `Somewhat = 1`, `No = 0`. **Overall score** = sum of all four.

**Instructions:**
- Read the Question and Answer carefully.
- For each criterion, choose one rating and give a short explanation ($\leq$ 30 words) in English.
- Output *only* one minified JSON object; no prose, no code fences.

**Output Schema (strict):**

```
{
 "clarity":{"rating":"Yes|Somewhat|No",
 "explanation":"..."},
 "helpfulness":{"rating":"Yes|Somewhat|No",
 "explanation":"..."},
 "accuracy":{"rating":"Yes|Somewhat|No",
 "explanation":"..."},
 "completeness":{"rating":"Yes|Somewhat|No",
 "explanation":"..."},
 "scores":{"clarity":int,"helpfulness":int,
 "accuracy":int,"completeness":int,"overall":int}
}
```

**Question:** question **Answer:** answer

Return the JSON now.

**Pairwise Comparison Prompt**

**Task:** Compare two answers to the same question and identify which one is better overall.
**Audience:** Internal evaluator. Use only the question, the two answers, and general world knowledge. No external browsing.
**Languages:** The question and answers may be in Hindi, Malayalam, or Kannada. Judge quality within the language of the answers, but write the explanation in concise English (maximum 40 words).

**Holistic Evaluation Criteria:** Consider clarity and fluency, helpfulness and relevance, factual accuracy or plausibility, completeness (no critical omissions), and safety (no harmful or misleading content). Prefer the answer that better serves the user overall. Do not reward hallucinated or unsafe content.

**Decision Rule:**
- Choose `A` or `B` if one answer is even slightly better overall.
- Choose `Not sure` only if both answers are essentially equal in quality or both are unintelligible or empty.

**Output Format (strict):** Return *only* a single JSON object with the exact keys:

```
{
  "winner": "A" | "B" | "Not sure",
  "explanation": "<concise justification □40 words>"
}
```

**Constraints:**
- `winner` must be exactly: A, B, or Not sure.
- `explanation` must be one sentence, ≤40 words, and must not quote A/B.
- No extra keys, no trailing commas, no additional text before or after the JSON.

**Question:** question
**Answer A:** answer_a
**Answer B:** answer_b

Return only the JSON now.

| Questions | Translation |
| --- | --- |
| എനിക്ക് PCOD യും തൈറോയ്ഡ് കൂടുതലുമുണ്ട്, എന്റെ കല്യാണം കഴിഞ്ഞിട്ട് ഇപ്പൊ 6മാസം ആയിട്ടുള്ളൂ, ഇതുവരെ ചികിത്സ ഒന്നും ചെയ്തിട്ടോ മരുന്ന് എടുക്കുകയോ ചെയ്തിട്ടില്ല. എനിക്ക് 1വർഷം ഒക്കെ കഴിഞ്ഞ് കുട്ടികൾ മതി എന്നാണ് തീരുമാനം. അതിനു വേണ്ടി ഇപ്പോമുതലെ ചികിത്സ തുടങ്ങണോ? കുട്ടിയെ പറ്റി ആലോചിക്കുന്ന സമയം മതിയാവുമോ ചികിത്സയും മരുന്നും ഒക്കെ? | I have PCOD and an overactive thyroid. It's been 6 months since my wedding, and I haven't had any treatment or taken any medication yet. I've decided that I'll have children after about a year. Should I start treatment now? Will I have enough time to think about the child, treatment, and medication? |
| മാനസിക ബുദ്ധിമുട്ട് നേരിടുന്ന എന്റെ ചേച്ചിയ്ക്ക് ക്ഷയരോഗം സ്ഥിരീകരിച്ചു. സ്വന്തം കാര്യം പോലും നോകാനുള്ള ബുദ്ധി വികസനം ഇല്ലാത്ത ചേച്ചിടെ കര്യങ്ങൾ എങ്ങനെയാണ് നോക്കേണ്ടത്? എന്തൊക്കെ മുൻകരുതലുകൾ ഞങ്ങൾ കാണണം? എങ്ങനെയാണ് ഇത് സുഖപ്പെടുത്തുക? | My sister, who is mentally challenged, has been diagnosed with Tuberculosis. How can care be provided for someone who doesn't have the cognitive ability to take care of herself? What precautions should be anticipated, and what does the treatment process involve? |
| गांव में अक्सर देखा जाता है कि लोग बिना डॉक्टर के सलाह के मेडिकल से ही दवाईयां लेते हैं, क्या यह रोग निदान का उचित तरीका है? इस विषय पर लोगों में जागरूकता लाने के लिए क्या कदम उठाए जा सकते हैं? | It is often seen in the village that people take medicines from pharmacies without a doctor's advice; is this a proper method of disease diagnosis? What steps can be taken to raise awareness among people on this issue? |
| क्या बच्चों को उनकी मनपसन्द चीज़े जो उनके स्वास्थ्य के लिए अच्छी नहीं हो उसके लिए मना करना उनके मानसिक स्वास्थ पर प्रभाव कर सकता है? किस तरह के खाद्य पदार्थों से उन्हें दूर रखना चाहिए और उसके बदले में क्या देना चाहिए जो बच्चों को भी पंसद आए । | Can prohibiting children from their favorite things that are not good for their health impact their mental health? What types of foods should they be kept away from, and what should be offered instead that children will also enjoy? |
| ನನ್ನ ತಂದೆ ತಾಯಿ ಇಬ್ಬರಿಗೂ ಶುಗರ್ ಇತ್ತು, ಅವರಿಂದ ನನಗೂ ಶುಗರ್ ಬರುತ್ತದೆಯೇ | Both my mother and my father had sugar (gloss: diabetes).Will I also get it from them? |
| ನಮ್ಮನೆ ಹತ್ರ ಇರೋ ಪ್ರಾಥಮಿಕ ಆರೋಗ್ಯ ಕೇಂದ್ರದಲ್ಲಿ ತೋರ್ಸೋಕೆ ಹೋದ್ರೆ ಎಲ್ಲಾದಕ್ಕೂ ದುಡ್ಡು ಕೇಳ್ತಾರೆ ನಾನ್ ಇವ್ರ್ ಮೇಲೆ ದೂರು ಕೊಡ್ಬೇಕು ಅಂದ್ರೆ ಏನ್ ಮಾಡ್ಬೇಕು? | If we seek treatment at the primary health center near my home, they ask for money. If I want to complain about this, what should I do? |

**Table 7: Additional user queries**